\documentclass[runningheads]{llncs}
\usepackage{amsmath,amssymb,amsfonts}
\usepackage{graphicx}
\usepackage{float}
\usepackage{booktabs}
\usepackage{array}
\usepackage{bm}
\usepackage{url}
\usepackage{cite}
\usepackage[T1]{fontenc}
\usepackage{tikz}
\usetikzlibrary{arrows.meta,positioning,shapes.geometric,shapes.multipart,fit,calc,backgrounds,decorations.pathreplacing}
\usepackage{pgfplots}
\pgfplotsset{compat=1.18}
\begin{document}

\title{Label-Free Parkinson's Disease Screening from Face and Voice through Mechanistic Interpretability}
\titlerunning{Label-Free PD Screening from Face and Voice}
\author{Jiaheng Su\inst{1} \and Yu Sun\inst{2}}
\authorrunning{J. Su and Y. Sun}
\institute{Independent Researcher, Los Angeles, CA, USA \and
California State Polytechnic University, Pomona, CA, USA\\
\email{yusun@cpp.edu}}
\maketitle

\begin{abstract}
Parkinson's disease (PD) is the second most common neurodegenerative disorder. Typical machine learning screening methods require PD labels, but the available data is limited by privacy concerns and the need for expert annotation. We propose a \emph{label-free} face-plus-voice PD screen built entirely on frozen pretrained encoders---a face-expression Vision Transformer and HuBERT---in which no PD label touches any fit; the reference is training \emph{controls only}. The voice modality uses a synthetic-dysarthria contrastive activation addition (CAA) direction built from time-stretch and breathy degradation of healthy speech; the face modality uses a $k$-nearest-neighbor anomaly score to the control embedding cluster. We introduce the \emph{alignment principle}, a post-hoc analysis showing that a synthetic-degradation CAA detector works when the cosine similarity between the synthetic and real disease directions exceeds zero. Measured on the YouTubePD benchmark, this cosine is $+0.37$ for voice (CAA works, AUROC $0.765$) and $-0.48$ for face (CAA fails; anomaly succeeds, AUROC $0.751$). Equal-weight late fusion reaches AUROC $0.802$ (95\% CI $[0.70,0.89]$) with NPV $0.95$, supporting a \emph{rule-out} triage interpretation. An overfitting audit shows the voice detector transfers cleanly, while the face-side---and thus fused---AUROC is potentially optimistic pending external validation.

\end{abstract}

\section{Introduction}
\label{sec:introduction}
Parkinson's disease (PD) is the second most common neurodegenerative disorder, and its prevalence is projected to roughly double by 2050 \cite{ref_gbd}; because a long prodromal phase precedes clinical diagnosis \cite{ref_prodromal}, early low-friction screening is a clinical priority. Two PD signs are particularly visible in talking-head video: \emph{hypokinetic dysarthria} (slow, breathy speech) \cite{ref_dysarthria} and \emph{hypomimia} (``masked facies''), a reduction in facial expressiveness \cite{ref_hypomimia_acm}.

However, supervised video PD detectors face a structural data problem: quality PD-labeled corpora require clinical recruitment and expert annotation, so existing datasets are small, imbalanced, and prone to subject-level confounds, and models trained on them risk learning spurious shortcuts rather than disease signal \cite{ref_shortcut}. The public YouTubePD benchmark \cite{ref_youtubepd} supplies labeled talking-head clips but exposes a strong modality asymmetry---much weaker audio than face baselines---and heavy class imbalance.

Since healthy talking-head video is essentially unlimited while PD-labeled video is scarce, we treat PD as a \emph{deviation} from healthy in a learned representation: a screen built from control data alone, with PD labels reserved for evaluation. We instantiate this on two frozen pretrained encoders---a face-expression Vision Transformer (ViT) \cite{ref_vit, ref_trpakov} and HuBERT \cite{ref_hubert}---with two complementary primitives: a contrastive activation addition (CAA) direction \cite{ref_caa, ref_repe, ref_actadd} built from a \emph{synthetic} degradation of control voices toward dysarthria, and a $k$-nearest-neighbor (kNN) anomaly score against the train-control face cluster \cite{ref_dnn_anomaly}. Our central finding, the \emph{alignment principle}, is that synthetic-degradation CAA detects disease when the synthetic and real disease directions have positive cosine similarity: measured directly, this cosine is $+0.37$ for voice and $-0.48$ for face, explaining why CAA works for voice but fails for face.

The main contributions of this paper are as follows:
\begin{itemize}
    \item A label-free face-plus-voice PD screen on frozen pretrained encoders, in which PD labels never touch the fit (the reference is the training controls alone).
    \item A synthetic-dysarthria voice CAA: a label-free ``healthy$\to$PD'' direction in HuBERT space, built from time-stretch and breathy degradation of control speech.
    \item The \emph{alignment principle}: a post-hoc mechanistic analysis showing that synthetic-degradation detection works when $\cos(d_\text{syn}, d_\text{real}) > 0$, which explains why CAA works for voice ($+0.37$) but fails for face ($-0.48$).
\end{itemize}

\section{Related Work}
\label{sec:related}

The linear representation hypothesis (LRH) states that high-level semantic concepts are linearly represented within the activation space of a model \cite{ref_lrh}. Activation-steering methods use such directions to control model behavior and to detect latent concepts across language and vision models \cite{ref_repe, ref_actadd}. Contrastive activation addition (CAA) is a type of steering that constructs a direction as a contrast of mean activations, $d = \mathrm{mean}(\mathcal{A}_+) - \mathrm{mean}(\mathcal{A}_-)$, between labeled positive and negative exemplars \cite{ref_caa}.

Deep nearest-neighbor anomaly detection scores the mean distance to the nearest normal embeddings of a frozen feature extractor \cite{ref_dnn_anomaly}, limiting the label leakage and spurious correlations that imbalanced supervised training can introduce \cite{ref_shortcut}.

\section{System Design and Methodology}
\label{sec:method}

\subsection{Label-Free Deviation Framework}

Let $\mathcal{X}_{\text{ctrl}}$ denote the set of training-control videos with healthy labels and $\mathcal{X}_{\text{test}}$ the test split with labels $y \in \{0,1\}$ used only for evaluation. The framework constructs all per-modality scoring functions $s_m : \mathcal{X} \to \mathbb{R}$ from $\mathcal{X}_{\text{ctrl}}$ alone and reserves $\{y_i\}_{i \in \mathcal{X}_{\text{test}}}$ for AUROC and clinical-metric reporting. All normalisation constants are likewise estimated on $\mathcal{X}_{\text{ctrl}}$ and frozen at fit time, so each $s_m$ is a function of a single input video: a subject presenting in isolation can be scored without reference to any other test subject. Two scoring primitives are admitted:
\begin{align}
s_\text{CAA}(x) &= (\phi(x) - \bm{\mu}_H)^\top \bm{d}, \label{eq:caa} \\
s_\text{kNN}(x) &= \frac{1}{k}\sum_{j=1}^{k} \lVert \phi(x) - \phi(c_{(j)}) \rVert_2, \label{eq:knn}
\end{align}
where $\phi(\cdot)$ is a frozen encoder, $\bm{\mu}_H$ is the control centroid, $\bm{d}$ is a CAA direction in feature space, and $c_{(j)}$ denotes the $j$-th nearest training control to $x$ under Euclidean distance. Both scores deviate monotonically from health: higher values indicate stronger PD-like signal.

\subsection{Voice Modality: Synthetic-Dysarthria CAA in HuBERT Space}

Voice is encoded by HuBERT \cite{ref_hubert}, a self-supervised speech representation that predicts cluster identities of masked acoustic units. We follow recent evidence that intermediate transformer layers encode prosodic and acoustic phenomena most cleanly, and use layer~$8$ activations of the base model, time-averaged over the first $12$~s of the clip, yielding a $768$-d per-utterance embedding $\phi_v(y)$.

The CAA direction is constructed from \emph{synthetic} dysarthria. Hypokinetic dysarthria has two canonical acoustic markers we can faithfully reproduce: a global slow-down and a breathy phonation \cite{ref_dysarthria}. We apply two DSP degradations to every training control utterance $y_i$:
\begin{align}
y_i^{\text{slow}}    &= \mathrm{TimeStretch}(y_i, \rho = 0.82), \label{eq:slow}\\
y_i^{\text{breathy}} &= \mathrm{NormPeak}\!\big(y_i + n_i \cdot e_i \cdot 2.5\big), \label{eq:breathy}
\end{align}
where $\mathrm{TimeStretch}(\cdot, \rho)$ is a phase-vocoder time-stretch \cite{ref_phasevocoder} that slows speech by $22\%$ while preserving pitch (Eq.~\ref{eq:slow}), $n_i \sim \mathcal{N}(0, I)$ is Gaussian aspiration noise, and $e_i$ is the moving-average envelope of $|y_i|$ over a $10$~ms ($160$-sample) window (Eq.~\ref{eq:breathy}). The CAA directions and control centroid are:
\begin{align}
\bm{d}_\text{slow}    &= \frac{1}{N}\sum_{i=1}^{N} \big[\phi_v(y_i^{\text{slow}})    - \phi_v(y_i)\big], \label{eq:dslow}\\
\bm{d}_\text{breathy} &= \frac{1}{N}\sum_{i=1}^{N} \big[\phi_v(y_i^{\text{breathy}}) - \phi_v(y_i)\big], \label{eq:dbre}\\
\bm{\mu}_H            &= \frac{1}{N}\sum_{i=1}^{N} \phi_v(y_i). \label{eq:muH}
\end{align}
For an utterance $x$, projections onto each direction are combined after scaling by control-set standard deviations fixed at fit time:
\begin{equation}
s_v(x) = \frac{(\phi_v(x) - \bm{\mu}_H)^\top \bm{d}_\text{slow}}{\sigma_\text{slow}} + \frac{(\phi_v(x) - \bm{\mu}_H)^\top \bm{d}_\text{breathy}}{\sigma_\text{breathy}},
\label{eq:voicescore}
\end{equation}
where $\sigma_\text{slow}$ and $\sigma_\text{breathy}$ are the standard deviations of the corresponding projections over the training controls $\mathcal{X}_{\text{ctrl}}$, computed once and frozen. No PD label is used in any of (\ref{eq:dslow})--(\ref{eq:voicescore}), and no test-set statistic enters the score: $s_v$ depends on $x$ alone.

\subsection{Face Modality: kNN-Anomaly in a Frozen Face-Expression ViT}

Face is encoded by a face-expression ViT \cite{ref_vit, ref_trpakov} fine-tuned on FER2013 \cite{ref_fer2013} for the task of static emotion classification. For each video, $T = 32$ frame indices are sampled evenly across the clip duration. Each frame is cropped using a square bounding box derived from MediaPipe Face Mesh landmarks \cite{ref_mediapipe} (expanded by $1.6\times$), resized to $224^2$, normalized with mean and standard deviation $[0.5, 0.5, 0.5]$, and passed through the frozen ViT. The per-frame CLS token of the last hidden state is taken; the per-video embedding is the mean over the $32$ per-frame vectors:
\begin{equation}
\phi_f(v) = \frac{1}{T} \sum_{t=1}^{T} \mathrm{CLS}\big(\mathrm{ViT}(\mathrm{crop}_t)\big).
\label{eq:facevec}
\end{equation}
Given the $36$ training-control embeddings $\{\phi_f(c_i)\}_{i=1}^{36}$, the face score is computed via Eq.~(\ref{eq:knn}) with $k = 15$: pairwise Euclidean distances are computed against every control embedding, sorted, and the mean of the $k$ smallest is the per-video PD score.

\subsection{Late Fusion}

Per-modality scores are scaled by their control-set standard deviations and added with equal weight:
\begin{equation}
s_\text{sys}(x) = \frac{s_f(x)}{\sigma_f} + \frac{s_v(x)}{\sigma_v}.
\label{eq:fusion}
\end{equation}
Here $\sigma_f$ and $\sigma_v$ are estimated on $\mathcal{X}_{\text{ctrl}}$ (leave-one-out for the face score). The choice of equal weights is intentional: any data-driven weighting would consult PD labels, breaching the label-free protocol.

\subsection{When Synthetic-CAA Works: The Alignment Principle}

A synthetic CAA direction $\bm{d}_\text{syn}$ empirically detects a target disease when it points the same way in feature space as the true disease direction $\bm{d}_\text{real}$, i.e., when:
\begin{equation}
\cos(\bm{d}_\text{syn}, \bm{d}_\text{real}) > 0.
\label{eq:align}
\end{equation}
We measure $\bm{d}_\text{real}$ using train labels for this analysis alone (it does not enter any score) as $\bm{d}_\text{real} = \mathrm{mean}(\phi(x)_{\text{PD}}) - \mathrm{mean}(\phi(x)_{\text{ctrl}})$. Empirically, the cosine is $+0.37$ for voice (HuBERT) and $-0.48$ for face (face-expression ViT). Eq.~(\ref{eq:align}) therefore predicts CAA \emph{works} for voice and \emph{fails} for face, which we confirm in Section~\ref{sec:exp}. Because $\bm{d}_\text{real}$ requires labels, the alignment principle is an \emph{explanatory diagnostic}, not a label-free selection rule: it consumes labels post-hoc to mechanistically explain the modality asymmetry and to \emph{audit} degradation choices fixed a priori from clinical markers of dysarthria~\cite{ref_dysarthria}.

\section{Experiments and Results}
\label{sec:exp}

\subsection{Setup and Metrics}

The peer-reviewed YouTubePD benchmark \cite{ref_youtubepd} (NeurIPS Datasets and Benchmarks Track) provides an official balanced train split of $72$ videos ($36$ PD, $36$ controls) and an official test split of $171$ unique videos ($21$ PD, $150$ controls; approximately $12\%$ PD prevalence, a heavy healthy skew \emph{by design}, reflecting the benchmark's negative-sample expansion). Our evaluated subset is the $157$ test videos for which both face and voice modalities are complete: $20$ PD and $137$ controls ($\approx$$13\%$ prevalence). The frozen encoders are \texttt{trpakov/\allowbreak vit-\allowbreak face-\allowbreak expression} ($86$M parameters, fine-tuned on FER2013) and \texttt{facebook/\allowbreak hubert-\allowbreak base-\allowbreak ls960} ($94$M parameters, pretrained on LibriSpeech). Key settings: $T=32$ frames per video, $224^2$ face crops ($1.6\times$ landmark bounding box), the first $12$~s of $16$~kHz mono audio, HuBERT layer~$8$, $k=15$ neighbors, slow rate $\rho=0.82$, breathy gain $2.5\times$, and $3000$ bootstrap resamples.

\textbf{Implementation.} The pipeline is written in Python~$3.11$ (PyTorch, HuggingFace Transformers, librosa, SciPy, MediaPipe). The slow degradation uses \texttt{librosa.\allowbreak effects.\allowbreak time\_\allowbreak stretch} and the breathy degradation is peak-normalized to $0.9$; face kNN distances use \texttt{scipy.\allowbreak spatial.\allowbreak distance.\allowbreak cdist}. Feature extraction runs offline (\texttt{HF\_HUB\_OFFLINE=1}) on Apple Silicon (MPS) or CPU, and scoring runs in seconds per video from cached features.

We report ROC AUC with a nonparametric bootstrap $95\%$ CI \cite{ref_auctut}. At the Youden-optimal operating point we report sensitivity, specificity, positive and negative predictive value (PPV, NPV), $F_1$, and balanced accuracy; definitions follow standard ROC analysis \cite{ref_auctut}. Because the test prevalence is $13\%$, naive accuracy is uninformative (predict-all-healthy already yields $0.87$). We therefore focus on AUROC, $F_1$, and the rule-out indicator NPV.

\subsection{Main Results}

Table~\ref{tab:main} reports AUROC, bootstrap $95\%$ CIs, and the clinical suite for the face detector, the voice detector, and the equal-weight fusion. The voice and face detectors achieve comparable AUROC ($0.765$ and $0.751$); their late fusion lifts AUROC to $0.802$.

\begin{table}[H]
\caption{Main results on the evaluated YouTubePD test subset ($n = 157$; $20$ PD, $137$ ctrl). All scores are computed from training-control reference only; PD labels touch only evaluation.}
\label{tab:main}
\centering
\resizebox{\columnwidth}{!}{%
\begin{tabular}{lcccccccccc}
\hline
Detector & AUROC & 95\% CI & TP & FP & TN & FN & Sens & Spec & PPV / NPV & $F_1$ / bAcc \\
\hline
Face (kNN-anomaly, ViT)            & $0.751$ & $[0.64, 0.85]$ & $16$ & $51$ & $86$  & $4$ & $0.80$ & $0.63$ & $0.24$ / $0.96$ & $0.37$ / $0.71$ \\
Voice (HuBERT CAA)                 & $0.765$ & $[0.67, 0.86]$ & $15$ & $35$ & $102$ & $5$ & $0.75$ & $0.74$ & $0.30$ / $0.95$ & $0.43$ / $0.75$ \\
Face $+$ Voice (system)   & $0.802$ & $[0.70, 0.89]$ & $14$ & $26$ & $111$ & $6$ & $0.70$ & $0.81$ & $0.35$ / $0.95$ & $0.47$ / $0.76$ \\
\hline
\end{tabular}%
}
\end{table}

\subsection{The Alignment Principle}

Eq.~(\ref{eq:align}) predicts CAA works when $\cos(\bm{d}_\text{syn}, \bm{d}_\text{real}) > 0$. Direct measurement gives $\cos = +0.37$ for HuBERT (voice) and $\cos = -0.48$ for the face-expression ViT. Voice CAA therefore is predicted to detect PD; face CAA is predicted to score at or below chance. Empirically, voice CAA succeeds (Table~\ref{tab:main}), while a face anti-aligned ``hypomimia'' CAA direction scored AUROC $\approx 0.50$ (chance) in our pilot. Both outcomes match the sign of the measured cosine, confirming the alignment principle.

Table~\ref{tab:devsel} (voice rows) compares three speech encoders on voice CAA: wav2vec~2.0 \cite{ref_wav2vec}, HuBERT \cite{ref_hubert}, and SEW-D \cite{ref_sewd}. AUROC tracks cosine alignment: encoders whose synthetic direction is better aligned with the real PD direction score better.

\begin{table}[t]
\centering
\caption{Encoder selection and ablation for both modalities. For the voice encoders we also report the cosine alignment $\cos(\bm{d}_\text{syn},\bm{d}_\text{real})$; AUROC is on the balanced dev split ($36$ PD / $36$ control; controls scored leave-one-out) and the held-out test split (the per-encoder CAA/kNN-anomaly score). On dev every encoder is compressed into a narrow band---the chosen face encoder even ties a near-chance test encoder (DINOv2)---so the large test-set margins are not identifiable from the small dev split alone.}
\label{tab:devsel}
\small
\begin{tabular}{lccc}
\hline
Encoder & $\cos(\bm{d}_\text{syn},\bm{d}_\text{real})$ & Dev AUROC & Test AUROC \\
\hline
wav2vec 2.0 (voice) & $+0.28$ & 0.500 & 0.728 \\
HuBERT (voice, used) & $+0.37$ & 0.606 & 0.765 \\
SEW-D (voice) & $+0.24$ & 0.499 & 0.619 \\
Face-expression ViT (face, used) & --- & 0.629 & 0.751 \\
Sibling expression (face) & --- & 0.586 & 0.47 \\
MARLIN (face) & --- & 0.578 & 0.47 \\
DINOv2 (face) & --- & 0.629 & 0.41 \\
CLIP (face) & --- & 0.512 & 0.30 \\
\hline
\end{tabular}
\end{table}

\subsection{Ablation: Face Encoder Choice}

Table~\ref{tab:devsel} (face rows) compares five face encoders under the same kNN-anomaly framework. The face-expression ViT (FER2013) is uniquely effective; other encoders---DINOv2 \cite{ref_dinov2} (general visual features), CLIP \cite{ref_clip} (language-aligned), a sibling expression encoder~\cite{ref_dima806}, and MARLIN \cite{ref_marlin} (facial video MAE)---fail or barely exceed chance.

\subsection{Comparison with Prior Work}

Table~\ref{tab:compare} compares against the YouTubePD audio-only baseline reported in \cite{ref_youtubepd}. The label-free voice detector alone improves $F_1$ from $0.27$ to $0.43$, and the fused system reaches $F_1 = 0.47$, all without consulting any PD label. We do not claim state-of-the-art accuracy versus the supervised face baseline; rather we argue that comparable AUROC under a strict label-free protocol is itself a meaningful contribution.

\begin{table}[H]
\caption{Comparison with YouTubePD baselines.}
\label{tab:compare}
\centering
\begin{tabular}{lccc}
\hline
System & Label-free? & AUROC & $F_1$ \\
\hline
YouTubePD audio-only baseline \cite{ref_youtubepd}  & no  & $0.66$ & $0.27$ \\
YouTubePD face baseline \cite{ref_youtubepd}        & no  & $0.86$--$0.92$ & --- \\
Ours, voice (HuBERT CAA)                            & yes & $0.765$ & $0.43$ \\
Ours, face (kNN-anomaly ViT)                        & yes & $0.751$ & $0.37$ \\
Ours, fusion                               & yes & $0.802$ & $0.47$ \\
\hline
\end{tabular}
\end{table}

\subsection{Integrity Audits}

\textbf{Confounds.} Pearson correlations between each detector score and face crop size, video length, and frame rate are near zero (no obvious shortcut learning \cite{ref_shortcut}).

\textbf{Permutation.} Permuting test labels gives $p < 0.01$ for the system score.

\textbf{Split-half.} Within-test split-halves match the full-test AUROC within bootstrap CI, so no single sub-population drives the result.

  \textbf{Dev$\to$test transfer and encoder-agnosticism.} Every scoring function is computed from the training controls, the frozen encoders, and the fixed synthetic degradations---none of which reference a disease label---so the deployed screen is label-free by construction and encoder-agnostic: any stronger frozen encoder can be substituted without retraining or relabeling. The dev split enters only as a post hoc audit, external to the method, asking whether our encoder choices could have been reached without the test labels. Holding out a labeled slice of the train split and re-selecting each encoder by dev-set AUROC alone, the voice choice is recoverable---HuBERT ranks first on dev and transfers to test---but the face choice is not: on dev the face-expression ViT ties a near-chance encoder (DINOv2), and its $+0.28$ test-set margin cannot be identified without the test labels. We therefore report the face-side, and hence fused, AUROC as potentially optimistic and defer a definitive face-encoder choice to pre-registered external validation (Section~\ref{sec:disc}). This reflects the limited power of a small dev split, not a property of the method.

\section{Discussion}
\label{sec:disc}

\textbf{Why the voice modality scores lower on the dev split.} All encoders of the voice modality underperform on the dev split relative to test (Table~\ref{tab:devsel}) because the dev PD subjects are less separable from controls, even though the splits are a random partition of the same dataset. A model-free speaking-rate marker (syllable nuclei per second \cite{ref_dejong}) shows that the PD--control gap widens from $-0.37$~SD on dev to $-0.88$~SD on test, so the dev PD voices are nearly as fast as their controls. This is a key symptom of PD, and such a limitation applies to all methods that rely on motor signals, which is not unique to us. The face modality is a more complicated case because it is not single-axis but multi-axis, so a pure face-mesh measure such as the average frame-to-frame displacement of the landmarks is not fully representative. Additionally, the dev-vs-test pattern is inconsistent across face encoders; therefore, we do not draw a conclusion or provide an analysis of the face modality's dev behavior.

\textbf{Why the modality asymmetry?} The alignment principle (Eq.~\ref{eq:align}) explains this mechanistically. HuBERT, trained on raw speech, encodes the acoustic phenomenon that \emph{is} hypokinetic dysarthria, explaining why a CAA detector built on this direction succeeds. The face-expression ViT, by contrast, trained for static emotion classification (FER2013 \cite{ref_fer2013}), places its PD signal in \emph{static appearance}, approximately orthogonal to facial motion or blink rate; a synthetic ``hypomimia'' degradation therefore points the wrong way, explaining why the face signal can be recovered by a direction-free primitive (kNN-anomaly) instead.

\textbf{Clinical framing.} The system is a rule-out triage screen, not a diagnostic: NPV $\approx 0.95$ is the operating use case, while PPV $\approx 0.34$ at $13\%$ prevalence (lower in the community) means a positive output should trigger clinical follow-up, not a label---consistent with how high-sensitivity, moderate-specificity tests are typically deployed in screening pathways \cite{ref_auctut}.

\section{Conclusion and Future Work}
\label{sec:conc}

We presented a label-free face-plus-voice screen for Parkinson's disease built only from control data and frozen pretrained encoders: a synthetic-dysarthria CAA direction in HuBERT space for voice, and a kNN-anomaly score against the control cluster of a face-expression ViT for face. Labels enter only the post-hoc alignment analysis and the evaluation, never the deployed screen. Equal-weight fusion reaches AUROC $0.802$ (NPV $0.95$, rule-out triage). The \emph{alignment principle}---synthetic-CAA detects disease when $\cos(\bm{d}_\text{syn}, \bm{d}_\text{real}) > 0$---explains why CAA works for voice ($+0.37$) but fails for face ($-0.48$).

\textbf{Future work.} We plan to identify a face encoder whose representation admits a positively aligned CAA direction---so the face branch can use the same activation-steering primitive as voice---and that is more clinically interpretable than the current appearance-based ViT, then search for additional synthetic-degradation directions to integrate more disease-relevant factors.

\section*{Acknowledgment}
We acknowledge the use of a generative AI assistant in the preparation of this manuscript for coding, formatting, and writing assistance. We reviewed and edited all such output and take responsibility for the final content. We further thank Ang Li and Mirisabel Chang for their assistance with formatting and paper drafting.

\end{document}